\documentclass[letterpaper, 10 pt, conference]{ieeeconf}  

\IEEEoverridecommandlockouts                              

\title{\LARGE \bf
A Multi-Sensor Fusion Approach for Rapid Orthoimage Generation in Large-Scale UAV Mapping
}

\author{Jialei He$^{1,*}$, Zhihao Zhan$^{2,*}$, Zhituo Tu$^{1}$, Xiang Zhu$^{2,\dag}$, and Jie Yuan$^{1,\dag}$
\thanks{$^{*}$ Jialei He and Zhihao Zhan contributed equally to this work.}%
\thanks{$^{\dag}$ Xiang Zhu (xzhu@topxgun.com) and Jie Yuan (yuanjie@nju.edu.cn) are the corresponding authors.}%
\thanks{$^{1}$School of Electronic Science and Engineering, Nanjing University, Nanjing, 210023, China.}%
\thanks{$^{2}$TopXGun Robotics, Nanjing, 211100, China.}
}

\usepackage[table]{xcolor} 
\usepackage{algorithm}
\usepackage{algpseudocode}
\usepackage{graphicx}
\usepackage{tikz}
\usepackage{overpic}
\usepackage{multirow}
\usepackage{subcaption} 
\usepackage{rotating} 
\usepackage{caption}
\usepackage{float}  
\usepackage{url}

\begin{document}

\maketitle
\thispagestyle{empty}
\pagestyle{empty}

\begin{abstract}

Rapid generation of large-scale orthoimages from Unmanned Aerial Vehicles (UAVs) has been a long-standing focus of research in the field of aerial mapping. A multi-sensor UAV system, integrating the Global Positioning System (GPS), Inertial Measurement Unit (IMU), 4D millimeter-wave radar and camera, can provide an effective solution to this problem. In this paper, we utilize multi-sensor data to overcome the limitations of conventional orthoimage generation methods in terms of temporal performance, system robustness, and geographic reference accuracy. A prior-pose-optimized feature matching method is introduced to enhance matching speed and accuracy, reducing the number of required features and providing precise references for the Structure from Motion (SfM) process. The proposed method exhibits robustness in low-texture scenes like farmlands, where feature matching is difficult. Experiments show that our approach achieves accurate feature matching and orthoimage generation in a short time. The proposed drone system effectively aids in farmland management.

\end{abstract}

\section{INTRODUCTION}

UAVs equipped with high-resolution cameras have revolutionized geospatial data acquisition~\cite{mollick2023geospatial} by enabling precise orthoimage generation. These orthoimages provide valuable insights for Geographic Information Systems (GIS)~\cite{baltsavias1996digital}, environmental monitoring, disaster assessment, and agriculture~\cite{sheng2003true, wu2020analysis, tiwari2020developing}.

Mature orthoimage generation methods mainly rely on the SfM framework~\cite{ullman1979interpretation}, which reconstructs 3D spatial information from image sequences. Implemented in commercial softwares Pix4DMapper~\cite{vallet2011photogrammetric} or Photoscan~\cite{verhoeven2011taking} and open-source projects such as OpenMVG~\cite{moulon2013global}, OpenDroneMap~\cite{OpenDroneMap} and Map2DFusion~\cite{bu2016map2dfusion}, these methods follow a common workflow: feature extraction, feature matching, camera pose estimation, sparse reconstruction, dense point cloud generation, Digital Surface Model (DSM) creation~\cite{zhang2005automatic}, and orthoimage generation.

In practical applications, image feature matching is computationally intensive, particularly with large datasets~\cite{pan2010region,hassaballah2016image}. To improve efficiency, researchers have proposed optimizations such as graphics processing unit acceleration~\cite{garcia2010k,sharma2016high,cornelis2008fast} and the use of Fast Library for Approximate Nearest Neighbors (FLANN)~\cite{vijayan2019flann,muja2012fast}. Furthermore, high-precision DSMs necessitate the densification of sparse point clouds, a process that also requires substantial computation and time.

Mainstream orthoimage generation methods predominantly rely on cameras as the primary sensor~\cite{yang2022fast}, with feature matching and dense point cloud generation dominating processing time and limiting processing speed. Our approach employs multi-sensor fusion for UAV mapping, enhancing precision and optimizing complex terrain processing to achieve better temporal performance.

This study proposes a multi-sensor fusion approach for rapid orthoimage generation in large-scale scenes. The approach addresses key aspects such as sensor calibration, terrain generation using 4D millimeter-wave radar, prior-pose-optimized feature matching, and orthoimage generation. Notably, by leveraging rough prior pose information from multiple sensors during feature matching, our method effectively improves matching speed and provides an efficient solution for UAV mapping in challenging terrains. The main contributions of this paper are as follows:

\begin{itemize}

\item A multi-sensor fusion UAV mapping system, which integrates data from various sensors to enhance the efficiency and accuracy of geospatial information acquisition.
\item An optimized feature matching process that leverages prior pose information to substantially accelerate matching speed, reduce the number of required feature points, and enhance matching accuracy, while maintaining robustness in challenging scenarios such as agricultural fields.
\item A novel method for orthoimage generation, ensuring that the resulting orthoimages are of high quality, geometrically corrected, and closely aligned with the true spatial configuration of the terrain.

\end{itemize}

\begin{figure*}[t]
    \centering
    \includegraphics[width=\textwidth]{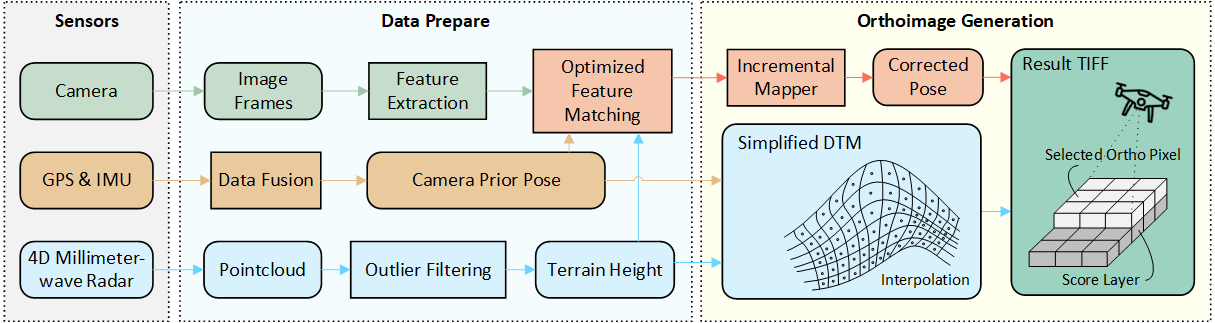}  
    \caption{Rapid orthoimage generation pipeline. Consecutive frames undergo feature extraction and matching using 4D radar terrain height and GPS-IMU pose data. Poses are corrected via SfM, and a simplified DTM is generated from the radar point cloud. The orthoimage is then produced by combining corrected poses and the DTM.}
    \label{figure1}
\end{figure*}

\section{Related Work}

Numerous studies~\cite{baltsavias1996digital,sheng2003true,tiwari2020developing,bu2016map2dfusion} have focused on generating orthoimages from UAV image sequences, aiming to improve accuracy, adaptability to terrains, and temporal performance. Traditional methods use 2D single-image correction with manual reference points, while modern approaches like SfM and SLAM reduce the reliance on ground control points, making UAV-based orthoimage generation more flexible. This also forms the core framework of our paper, which extends this approach by integrating multi-sensor data.

\subsection{2D Image Correction Approach} 

The 2D single image correction approach first performs ortho-rectification on the UAV images and then integrates them using 2D image stitching techniques. The foundational study by R. Szeliski in 2007~\cite{szeliski2007image} provides a comprehensive discussion of the early processes involved in this approach, which covers effective direct (pixel-based) and feature-based alignment algorithms, and describes blending algorithms used to produce seamless mosaics. Recent works~\cite{adel2014image,wang2020review} have also covered the latest methods within this category. The advantage of this method is its simplicity and directness. However, the separation of ortho-rectification and stitching in this approach makes it difficult to achieve optimal matching within a unified framework. Since this method relies solely on 2D image information, it struggles to handle complex scenes such as terrain variations, shadows, and occlusions. As a result, while it performs well in small scenes or simple image sequences, it tends to yield larger matching errors in large-scale scenes or when dealing with low-overlap images.

\subsection{SfM and SLAM-based Approach} 

The SfM and SLAM-based approaches have become popular research topics in recent years. SfM-based methods~\cite{moulon2013global,bu2016map2dfusion,schonberger2016structure,pan2024global} excel in 3D reconstruction accuracy and orthoimage quality but suffer from lower computational efficiency, making them more suitable for offline tasks. While SLAM-based methods~\cite{davison2007monoslam,wang2019terrainfusion} demonstrate superior real-time performance and can generate orthoimages in a relatively short time, though with lower accuracy compared to SfM methods.

As a fundamental task in computer vision, SfM plays a crucial role in various applications, such as 3D reconstruction~\cite{snavely2006photo,kerbl20233d}, novel view synthesis~\cite{kerbl20233d,mildenhall2021nerf} and orthoimage generation~\cite{wang2014automated,lv2024fast}. SfM typically involves two main approaches: incremental and global methods, which differ in how they handle the reconstruction and optimization processes. Process of incremental methods[28] is more commonly used in most open-source projects and commercial software due to their ability to process data progressively and optimize computational resources. However, considering the cumulative error issue inherent in incremental methods, many studies~\cite{schonberger2016structure,wu20153d,lao2018rolling} have enhanced reconstruction efficiency, accuracy, and robustness by modifying matching strategies, introducing scene augmentation, constraining camera pose estimation, and other approaches. Nevertheless, as the number of cameras increases, incremental methods encounter two primary challenges. First, feature matching across a large number of images requires substantial computational resources and can reduce efficiency. Second, performing global feature matching increases the likelihood of mismatches. In this study, we tackle these issues by implementing an optimized feature matching process that utilizes prior pose information.

Global methods, such as HSfM~\cite{cui2017hsfm}, require high accuracy in feature matching. In addition to incremental and global methods, Hierarchical SfM~\cite{lynen2013robust} employs an agglomerative clustering strategy to construct a binary cluster tree, but it cannot guarantee sufficient coverage of all scene structures by the selected images, making it less suitable for UAV-based aerial imagery.

\subsection{Multi-sensor Fusion Approach} 

Multi-sensor fusion~\cite{lynen2013robust,falquez2016inertial,hinzmann2018mapping} approaches are now widely used in SLAM, where the integration of IMU and GPS data enhances positioning accuracy. Recent work~\cite{he2025ligo} has achieved high-precision global positioning and real-time mapping by integrating data from LiDAR, IMU, and GNSS. Similarly, in orthoimage generation tasks~\cite{bu2016map2dfusion,wang2019terrainfusion,han2019automated,chen2020densefusion}, IMU and GPS can also be used to assist SfM systems in fully perceiving environmental information. Radar~\cite{hong2021radar,holder2019real} has been proven to be robust under various lighting conditions and adverse weather. The 4D millimeter-wave radar~\cite{han20234d}, by measuring the height of targets, enhances the ability to detect obstacles and terrain changes, providing more comprehensive environmental information. As a result, it is highly suitable for orthoimage generation in complex terrains and has been integrated into our multi-sensor fusion system.

\section{METHOD}

The main idea of this paper is to employ a multi-sensor fusion approach to build an orthoimage generation system adapts to complex terrains. Our primary goal is to maximize efficiency while ensuring the accuracy and robustness. The paper outlines the overall system workflow (Section III-A), multi-sensor synchronization (Section III-B), terrain map generation from radar data (Section III-C), prior-pose-optimized feature matching (Section III-D), and the orthoimage generation algorithm (Section III-E).

\subsection{System Overview} 

Fig.~\ref{figure1} illustrates the overall system workflow. During the drone flight, the camera continuously captures overlapping images with rapid feature extraction. Simultaneously, the 4D millimeter-wave radar collects point clouds to construct a terrain map and provides the drone's current altitude. The IMU and GPS supply the drone's position and orientation, establishing a rough prior pose for the camera and radar. Based on the extracted features, ground height, and prior pose, an optimized feature matching strategy is applied. The matching results are then fed into COLMAP's sparse reconstruction pipeline~\cite{schonberger2016structure} to recover precise camera extrinsic parameters. Finally, the terrain map are generated by the millimeter-wave radar, which serves as the Digital Terrain Model (DTM) for orthoimage generation.

\subsection{Multi-sensor synchronization} 

In order to improve the system's positioning accuracy and stability, we choose to fuse data from GPS and IMU sensors to achieve accurate localization during the UAV’s flight. The fusion of GPS and IMU data is achieved through the Extended Kalman Filter (EKF)~\cite{balamuruganky_ekf_2021} method, which combines the geographic positioning information provided by GPS and the dynamic information provided by IMU. The prediction-update cycle of the EKF effectively enhances the localization accuracy. The state vector we define is as follows:

\begin{equation} \label{eq1}
\mathbf{x}_k = \left[ \mathbf{p}_k^T, \mathbf{v}_k^T, \mathbf{q}_k^T, \mathbf{b}_k^T \right]^T,
\end{equation}

\noindent$\mathbf{p}_k$, $\mathbf{v}_k$, $\mathbf{q}_k$, and $\mathbf{b}_k$ represent the position vector, velocity vector, attitude quaternion, and IMU bias vector at time step $k$, respectively.

In the prediction step, the state of the system is propagated on the basis of the previous state and the IMU measurements. In the update step, the predicted state is corrected using the GPS measurements. 

With the covariance matrix of the state estimate $\mathbf{P}_k$, the Jacobian matrix of the measurement function $h(\cdot)$ represented as $\mathbf{H}$, the measurement noise covariance matrix $\mathbf{R}$, the Kalman gain $\mathbf{K}_k$ is computed to minimize the estimation error:
\begin{equation} \label{eq2}
\mathbf{K}_k=\mathbf{P}_k \mathbf{H}^T\left(\mathbf{H} \mathbf{P}_k \mathbf{H}^T+\mathbf{R}\right)^{-1}.
\end{equation}

Finally, the state vector is updated as follows:
\begin{equation} \label{eq3}
\mathbf{x}_k = \mathbf{x}_k + \mathbf{K}_k \left( \mathbf{z}_k - h(\mathbf{x}_k) \right).
\end{equation}

The iterative prediction-update cycle refines the UAV's state estimate at each time step, improving localization accuracy. The UAV's rotation matrix $\mathbf{R}_k$ from $\mathbf{q}_k$ and position vector $\mathbf{p}_k$ are used to derive the homogeneous transformation matrix $\mathbf{T}_B^W$. The 4D millimeter-wave radar and camera, fixed on the UAV, have extrinsic parameters $\mathbf{T}_R^B$ and $\mathbf{T}_C^B$, with which the real-time extrinsic parameters $\mathbf{T}_R^W=\mathbf{T}_B^W \cdot \mathbf{T}_R^B$ and $\mathbf{T}_C^W=\mathbf{T}_B^W \cdot \mathbf{T}_C^B$ are obtained. Synchronization of GPS, IMU, radar, and camera ensures all UAV data is unified in the global coordinate system, facilitating subsequent processing and analysis.

\begin{figure}[t]
    \centering
    \begin{overpic}[width=\columnwidth]{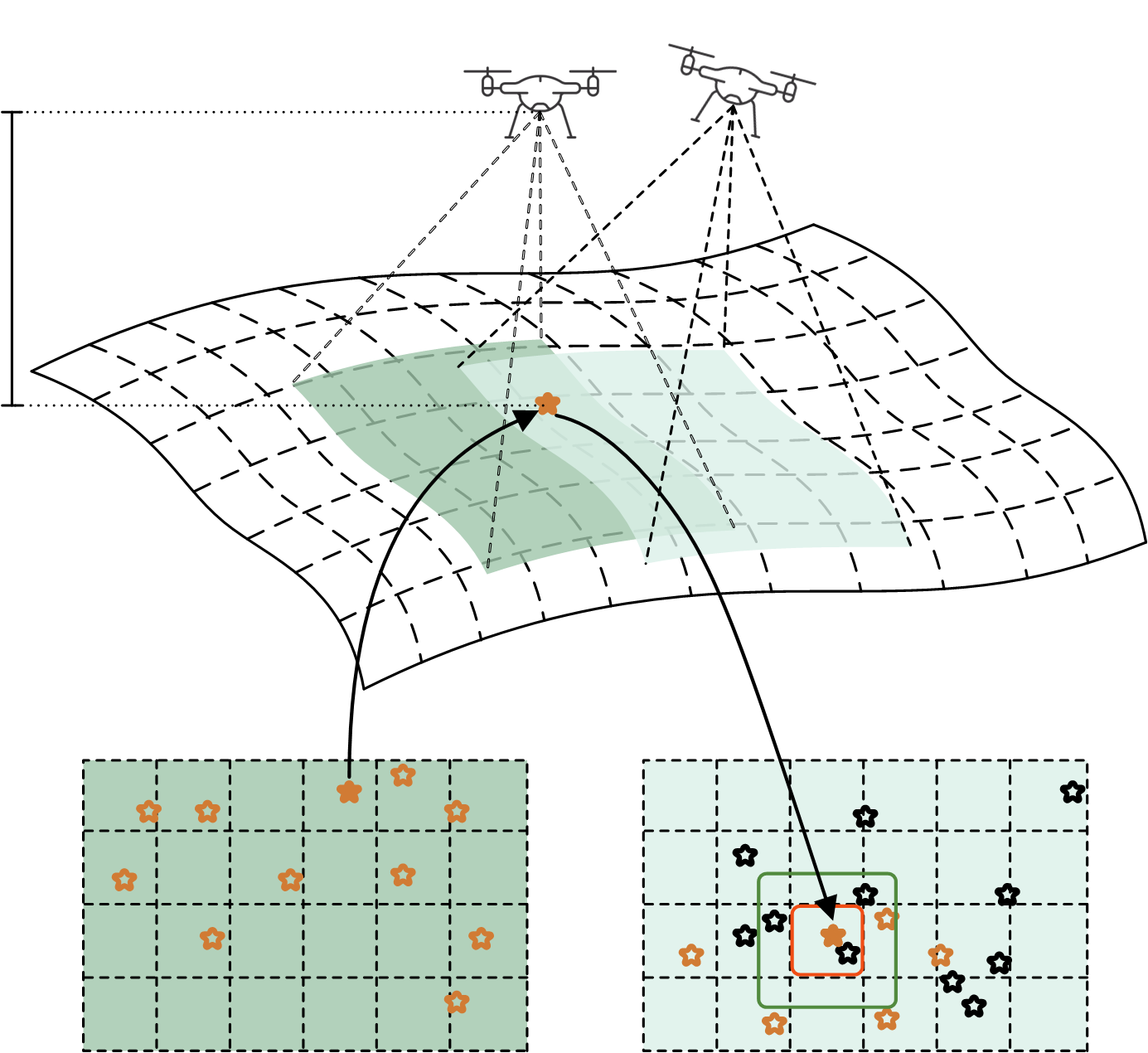}
        \put(3,68){\color{black} $\mathbf{Z}_W$}
        \put(7,28){\color{black} imageA}
        \put(56,28){\color{black} imageB}
    \end{overpic}
    \caption{Schematic of the prior-pose-optimized feature matching method. Scene height $\mathbf{Z_{W}}$ from radar, along with camera parameters and prior pose, is used to back-project feature points from image A into 3D coordinates and then project them into image B's pixel coordinate system. The pixel plane is divided into blocks for matching, with the search area enlarged due to initial pose uncertainty.}
    \label{figure2}
\end{figure}

\subsection{Terrain Map From Radar Points}

For the 4D millimeter-wave radar point clouds, preprocessing is first required. The terrain point $\mathbf{P}_R$ detected in the radar coordinate system are transformed into the UAV's world coordinate system using the real-time extrinsic parameters $\mathbf{T}_R^W$ of the millimeter-wave radar. This gives the position of the point in the world coordinate system as $\mathbf{P}_W=\mathbf{T}_R^W \cdot \mathbf{P}_R$.

The obtained point clouds undergo voxel downsampling to reduce the point density. Outlier points are then removed using statistical analysis methods. After preprocessing, the point clouds are divided into grids, with each grid storing a set of points. Within each grid, the points are sorted by height, and the point at the specified percentile is selected as a control point. If a grid contains no points, it is marked as invalid, and the interpolation method~\cite{foley1987interpolation} is applied. The processed control points are then used to estimate terrain height in accordance with their 2D coordinates, generating a simplified terrain map. This map is subsequently utilized for feature matching and orthoimage generation.

Compared to LiDAR, the 4D millimeter-wave radar enhances system robustness while further reducing hardware costs. In the task of generating outdoor large-scale terrain maps, especially in foggy or dusty environments, the reliability of the point cloud is more crucial than its density.

\subsection{Prior-pose-optimized Feature Matching}

Feature matching in orthoimage generation is a necessary but time-consuming step. Traditional methods use FLANN~\cite{vijayan2019flann} instead of brute-force matching~\cite{garcia2010k} to reduce computational complexity from $\mathcal{O}(n^2)$ to $\mathcal{O}(\log n)$ or $O(n \log n)$. 
To further improve speed and accuracy, we propose a novel feature matching approach that integrates multi-sensor data (GPS, IMU, and millimeter-wave radar) to align overlapping UAV images in the same pixel coordinate system. This method is compatible with any feature type and matching method. To enhance system efficiency, we aim to extract minimal yet precise feature points. We prioritize integrating optimization method into the brute-force matching scheme and testing it with the FLANN approach.

The prior-pose-optimized feature matching method in Fig.~\ref{figure2} utilizes the extrinsic parameters $\mathbf{T}_C^W$ and intrinsic parameters $\mathbf{K}$ of the camera to transform coordinates for matching images. For two overlapping images taken by the UAV-mounted camera, keypoint $\mathbf{P_{PA}} = \left( x_{PA}, y_{PA} \right)$ in image A are transformed into camera coordinates as $\mathbf{P_{CHA}} = \mathbf{K}^{-1} \left[ \begin{array}{c}
x_{PA} \\
y_{PA} \\
1
\end{array} \right]$ and scaled by terrain height $Z_W$ to obtain the actual coordinates $\mathbf{P}_{CA}$ in the camera coordinate system. Given that the camera is shooting almost vertically down when performing orthoimage missions, and the ground altitude is a low-frequency variation, we use the altitude of the aircraft $Z_W$ to the ground provided by the millimeter-wave radar point cloud to approximate the depth of the 3D point in the camera coordinate system. This point is then transformed to the world coordinate system using $\mathbf{T}_{CA}^W$, resulting in $\mathbf{P}_{WA} = \mathbf{T}_{CA}^W \cdot \mathbf{P}_{CA}$. Then, $\mathbf{P}_{WA}$ is converted into the pixel coordinates in image B with $\mathbf{T}_{CB}^W$:
\begin{equation} \label{eq4}
\mathbf{P}_{pAtoB}=\frac{\mathbf{K} \cdot \mathbf{{T}_{CB}^W}^{-1} \cdot \mathbf{P}_{WA}}{\left(\mathbf{{T}_{CB}^W}^{-1} \cdot \mathbf{P}_{WA}\right)_z},
\end{equation}

\noindent where $\left(\mathbf{{T}_{CB}^W}^{-1} \cdot \mathbf{P}_{WA}\right)_z$ is the $z$-component for normalizing the coordinates.

After unifying keypoints from both images into the pixel coordinate system of image B, the coordinate system is divided into square regions with side lengths of $blocksize$ as in Fig.~\ref{figure2}. Each block stores keypoints from both images and computes descriptor distances to select the best match. Due to pose inaccuracies, the search area for feature points in image B is expanded from $blocksize \times blocksize$ region to $(blocksize + 2 \times padding) \times (blocksize + 2 \times padding)$. When using the brute-force matching method, the computational complexity becomes $\mathcal{O}(n \cdot (blocksize + 2 \times padding)^2)$. The computational load is also significantly reduced when the FLANN method is used. Additionally, by limiting the number of feature points in each block, we can homogenize feature distribution, preventing over-concentration in texture-rich areas and reducing redundancy. 

This approach reduces the search space for feature matching by predicting feature point locations in the next frame, thereby lowering the matching complexity and mismatch rate. Consequently, fewer feature points are needed for the SfM process, significantly improving the speed of mapping algorithm.

\subsection{Orthoimage Generation}

Images are matched sequentially during the UAV's flight, and the matched pairs are subsequently fed into the SfM reconstruction process, where precise camera extrinsics are computed and aligned with the geodetic coordinate system. 

\begin{algorithm}[t]
\caption{Orthoimage Generation}\label{alg:ortho}
\begin{algorithmic}[1]
\Function{GenerateOrthoimage}{IMAGE, CAMERA EXTRINSIC and INTRINSICS, TERRAIN DATA}
    \State Initialize orthoimage and score layer
    \State Apply Illumination Normalization to $images$ 
    \For{each frame $f$ in frames}
        \State $image \gets$ ReadImage($f.img\_name$)
        \State Apply Gaussian Blur to $image$
        \State Compute affected grid range for $f$
        \For{each grid cell $(x, y)$ in affected range}
            \State $point \gets$ GetWorldPoint($x, y$)
            \State $z \gets$ QueryTerrainHeight($point$)
            \State $point3D \gets (point.x, point.y, z)$
            \State $score \gets$ ComputeScore($point3D, f.pos$)
            \If{$score >$ score\_layer[$x, y$]}
                \State Project $point3D$ to image $(u, v)$
                \If{$u, v$ are valid}
                    \State $pixel \gets$ GetPixel($image, u, v$)
                    \State $weight \gets$ ComputeWeight($score$)  
                    \State $pixel \gets$ BlendPixel($pixel, weight$)
                    \State Update orthoimage pixel with $pixel$
                    \State Update score\_layer[$x, y$] with $score$
                \EndIf
            \EndIf
        \EndFor
    \EndFor
\EndFunction
\end{algorithmic}
\end{algorithm}

Orthoimage generation typically involves creating a DSM from SfM-derived 3D point clouds. However, given the robustness and realism of 4D millimeter-wave radar point clouds, we opt to either fuse the two point clouds or directly use the simplified terrain map from the radar point cloud as an alternative.

Our orthoimage generation method, shown in Algorithm~\ref{alg:ortho}, adopts a backward projection approach from~\cite{hinzmann2018mapping}, but with customized terrain data management and viewpoint scoring mechanism. Additionally, we incorporate pixel value fusion processing to ensure the continuity and consistency of the generated orthoimage. To prevent aliasing and reduce stitching artifacts, Gaussian Blur is applied to smooth high-frequency details in the original images, following the Nyquist sampling theorem. 

\begin{table*}[t]
\caption{Quantitative comparison before and after optimization. The unit of time is second.}
\label{table2}
\begin{center}
\begin{tabular}{c|c c|c c c c|c}
\hline
Scene                 & \multicolumn{2}{c|}{Matching}    & Extraction Time        & Matching Time               & Mapper Time                & Total Time & 3D points               \\ \hline
\multirow{5}{*}{Road} &\multirow{3}{*}{BF}    & Opt.     & \multirow{5}{*}{4.449} & \cellcolor{red!15}4.347     & \cellcolor{red!15}23.756   & 32.552     & \cellcolor{red!15}15082 \\
                      &                       & Opt.+Hom.&                        & 4.617                       & 24.958                     & 34.024     & 9839                    \\
                      &                       & None     &                        & 138.907                     & 26.712                     & 170.068    & 14891                   \\ \cline{2-3} 
                      &\multirow{2}{*}{FLANN} & Opt.     &                        &  6.982                      & 23.374                     & 34.805     & \cellcolor{orange!10}15081 \\
                      &                       & None     &                        & 39.954                      & 26.466                     & 70.869     & 14831                   \\  \hline
\multirow{5}{*}{Hill} &\multirow{3}{*}{BF}    & Opt.     & \multirow{5}{*}{9.063} & \cellcolor{red!15}10.461    & 90.423                     & 109.947    & \cellcolor{red!15}32577 \\
                      &                       & Opt.+Hom.&                        & 11.346                      &  \cellcolor{red!15}63.857  & 84.266     & 21226                   \\
                      &                       & None     &                        & 274.045                     & 87.364                     & 370.472    & 31965                   \\ \cline{2-3}
                      &\multirow{2}{*}{FLANN} & Opt.     &                        & 16.866                      & 88.976                     & 114.905    &  \cellcolor{orange!10}32556 \\
                      &                       & None     &                        & 84.621                      & 93.230                     & 186.914    & 31602                   \\ \hline
\end{tabular}
\end{center}
\end{table*}

In Algorithm~\ref{alg:ortho}, the ground sampling distance (GSD) and camera's field of view (FOV) boundaries are determined and calculated to define each frame's affected area in the orthoimage, based on which the orthoimage size is then initialized along with a score map. Ranging from -1 (completely upward) to 1 (directly downward), the viewpoint score is based on the angle between the camera's line of sight and the vertical direction. A score closer to 1 indicates a better viewpoint. Specifically:

\begin{equation} \label{eq5}
score = \mathbf{v} \cdot \mathbf{u} = \frac{z - z_c}{\sqrt{(x - x_c)^2 + (y - y_c)^2 + (z - z_c)^2}},
\end{equation}

\noindent where $\mathbf{v}$ and $\mathbf{u}$ represent unit vectors for the line of sight and vertical directions. $(x, y, z)$ and $(x_c, y_c, z_c)$ are the world coordinates of the physical point and camera.

\section{EXPERIMENTS}

Considering the lack of public datasets with 4D radar point clouds, UAV images, and real-time pose information, we use our own collected data for experiments. We evaluate feature matching improvements and system robustness across scenarios, and generate orthoimages to validate the system's integrity and robustness.

All experiments are performed on a desktop PC running a Linux system, equipped with 2 Intel Xeon E5-2696 v2 @ 2.50GHz CPUs, 62 GiB of RAM (approximately 64 GB), and an NVIDIA Tesla M40 24GB GPU.

\subsection{Data Preparation}

The experimental data were collected using a TopXGun FP700 Agriculture Drone equipped with an IMX577 camera configured for infinite focal length, a UM982 RTK module, a Bosch BMI088 IMU, and a Mindcruise A1 radar. Two datasets from a road and a hillside scene of farmland were used (Table~\ref{table1}). The images have high similarity, no distinct markers, and a resolution of ${1920 \times 1080}$. The drone flew at $10 m/s$ during data collection. H-Max and H-Min in Table~\ref{table1} are the ground elevation heights.

\begin{table}[h]
\caption{Data used in experiments.}
\label{table1}
\begin{center}
\begin{tabular}{c|c c c c}
\hline
Scene & Frames & H-Max(m) & H-Min(m) & Path Length(m) \\ \hline
Road & 84 & 17.57 & 9.68 & 694.69 \\
Hill & 222 & 33.32 & 19.61 & 1871.12 \\
\hline
\end{tabular}
\end{center}
\end{table}

\begin{figure}[h]
    \centering
    \begin{subfigure}[h]{0.77\columnwidth}
        \centering
        \includegraphics[width=\linewidth]{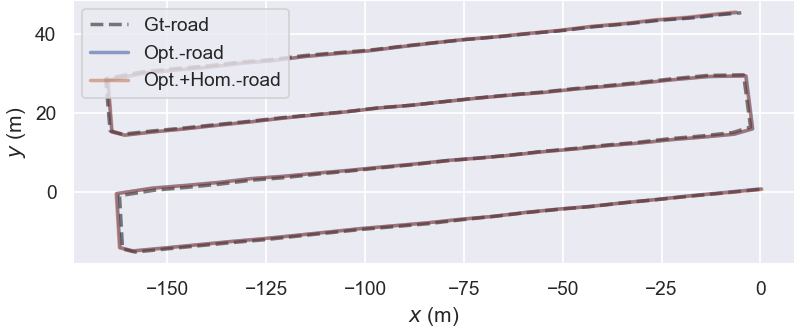}
        \caption{Road}
        \label{figure4a}
    \end{subfigure}
    \begin{subfigure}[h]{0.77\columnwidth}
        \centering
        \includegraphics[width=\linewidth]{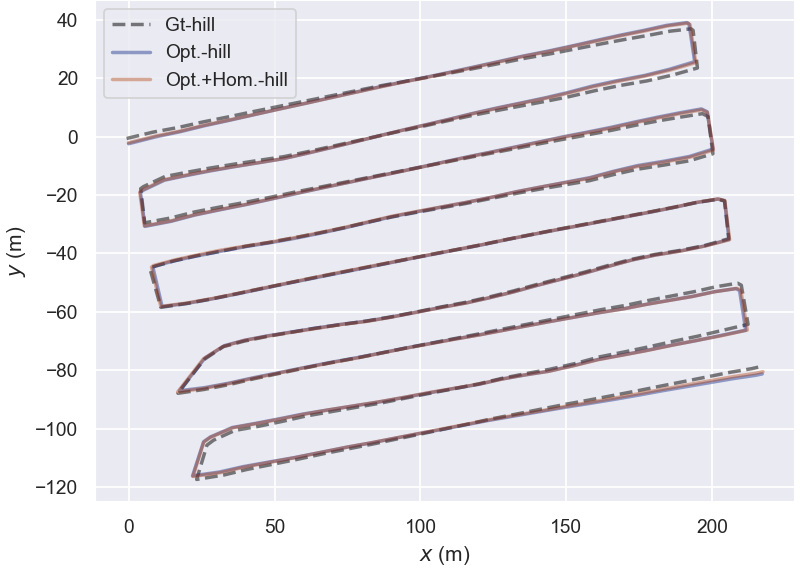}
        \caption{Hill}
        \label{figure4b}
    \end{subfigure}
    \caption{Trajectories evaluation with evo}
    \label{figure4}
\end{figure}

\subsection{Performance of Feature Matching}

We implemented the optimized feature matching process in COLMAP's ASIFT algorithm~\cite{schonberger2016structure}, and optimize both brute-force and FLANN matching. During optimization, $blocksize$ was set to $120$, and $padding$ to $blocksize/4$. The maximum number of feature points extracted is set to 2000, and the sequential matcher's overlap parameter was set to 4 for the road scene and 5 for the hill scene. Feature matching and mapper was performed on the CPU, while feature extraction on the GPU. Data statistics before and after optimization are shown in Table~\ref{table2}. 'Opt.' denotes optimization, 'Opt.+Hom.' indicates optimization with feature homogenization (retaining up to 4 pairs per block), and 'None' refers to the original method without optimization.

\begin{figure*}[t]
    \centering
    \begin{subfigure}[t]{0.9\textwidth}
        \centering
        \begin{overpic}[width=\linewidth]{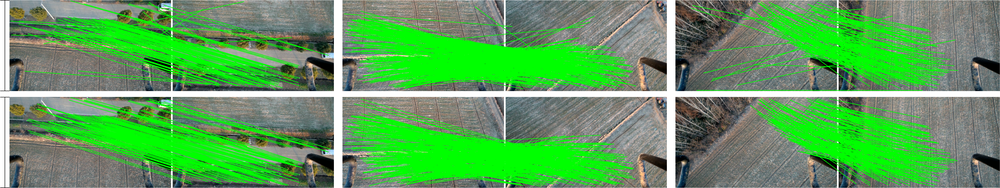}
            \put(-2, 3){\rotatebox{90}{\textbf{Opt.}}}
            \put(-2, 11){\rotatebox{90}{\textbf{None}}}
        \end{overpic}
        \caption{Road}
        \label{figure3a}
    \end{subfigure}
    \vfill
    \vspace{5pt} 
    \begin{subfigure}[t]{0.9\textwidth}
        \centering
        \begin{overpic}[width=\linewidth]{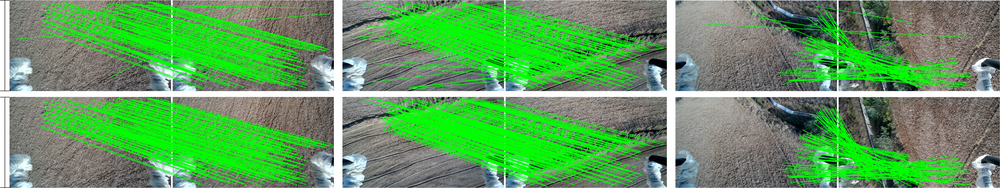}
            \put(-2, 3){\rotatebox{90}{\textbf{Opt.}}}
            \put(-2, 11){\rotatebox{90}{\textbf{None}}}
        \end{overpic}
        \caption{Hill}
        \label{figure3b}
    \end{subfigure}
    \caption{Qualitative comparison before and after optimization on both scenes}
    \label{figure3}
\end{figure*}

With high similarity and no distinct marker, data collected from farmlands pose significant challenges for conventional feature matching methods. The optimized matching achieves a 25× speedup in brute-force and a 5× speedup in FLANN compared to the original methods, while maintaining reconstruction quality. Optimization in both schemes reduces mapper time, recovers more reliable 3D points, and lowers the mismatch rate. Fig.~\ref{figure3} visualizes a significant reduction in mismatches before and after optimization using brute-force matching. The lower-left and lower-right corners, affected by the camera's mounting position, are masked out during feature extraction.

\begin{figure}[h]
    \centering
    \begin{subfigure}[h]{0.88\columnwidth}
        \centering
        \includegraphics[width=\linewidth]{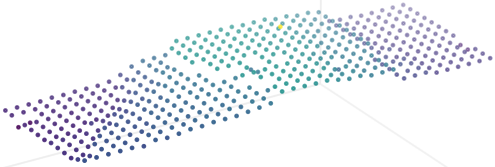}
        \caption{Road}
        \label{figure5a}
    \end{subfigure}
    \begin{subfigure}[h]{0.88\columnwidth}
        \centering
        \includegraphics[width=\linewidth]{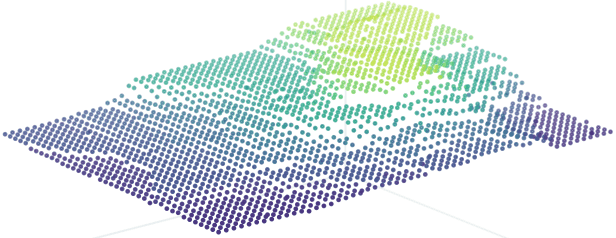}
        \caption{Hill}
        \label{figure5b}
    \end{subfigure}
    \caption{Simplified terrain map for orthoimages}
    \label{figure5}
\end{figure}

\begin{figure*}[h]
    \centering
    \begin{subfigure}[h]{0.28\textwidth}
        \centering
        \begin{overpic}[width=\linewidth]{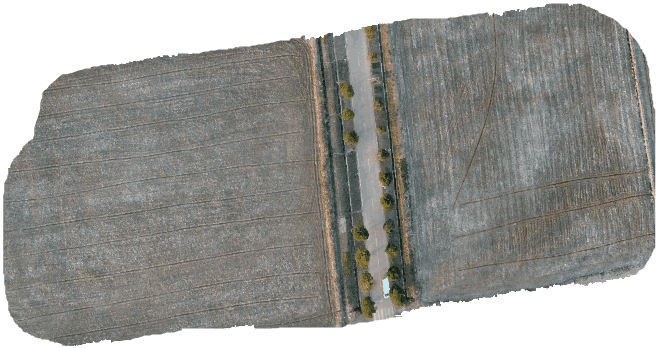}
            \put(-10, 12){\rotatebox{90}{\textbf{road}}}
        \end{overpic}
        \label{figure6a}
    \end{subfigure}
    \hspace{0.01\textwidth}
    \begin{subfigure}[h]{0.28\textwidth}
        \centering
        \includegraphics[width=\linewidth]{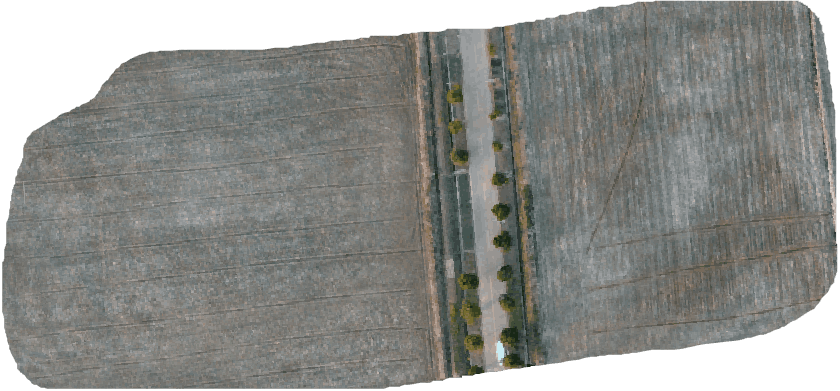}
        \label{figure6b}
    \end{subfigure}
    \hspace{0.01\textwidth}
    \begin{subfigure}[h]{0.28\textwidth}
        \centering
        \includegraphics[width=\linewidth]{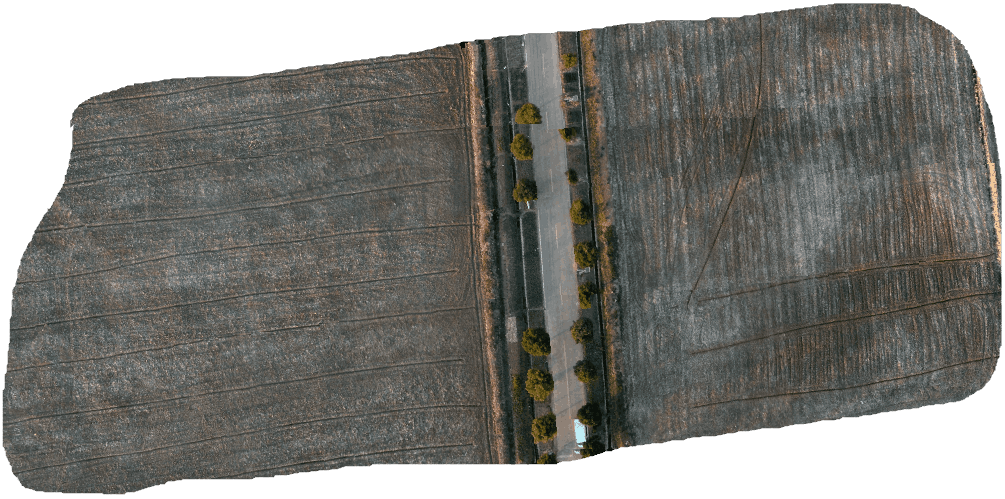}
        \label{figure6c}
    \end{subfigure}
    \begin{subfigure}[h]{0.28\textwidth}
        \centering
        \begin{overpic}[width=\linewidth]{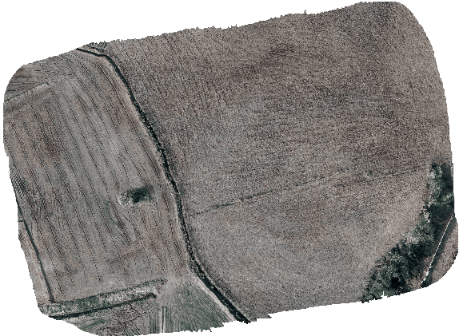}
            \put(-10, 28){\rotatebox{90}{\textbf{hill}}}
        \end{overpic}
        \caption{WebODM~\cite{OpenDroneMap}}
        \label{figure6d}
    \end{subfigure}
    \hspace{0.01\textwidth}
    \begin{subfigure}[h]{0.28\textwidth}
        \centering
        \includegraphics[width=\linewidth]{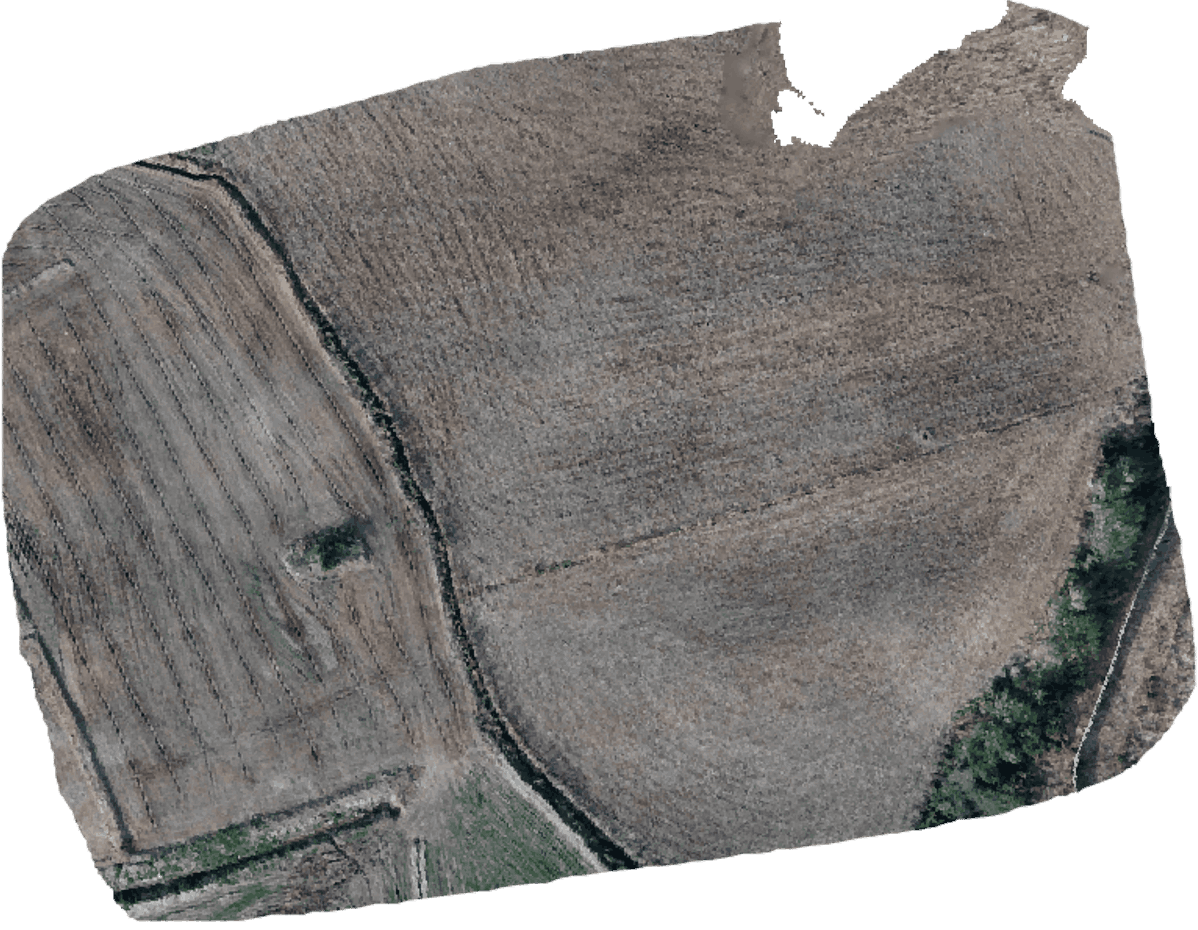}
        \caption{Photoscan~\cite{verhoeven2011taking}}
        \label{figure6e}
    \end{subfigure}
    \hspace{0.01\textwidth}
    \begin{subfigure}[h]{0.28\textwidth}
        \centering
        \includegraphics[width=\linewidth]{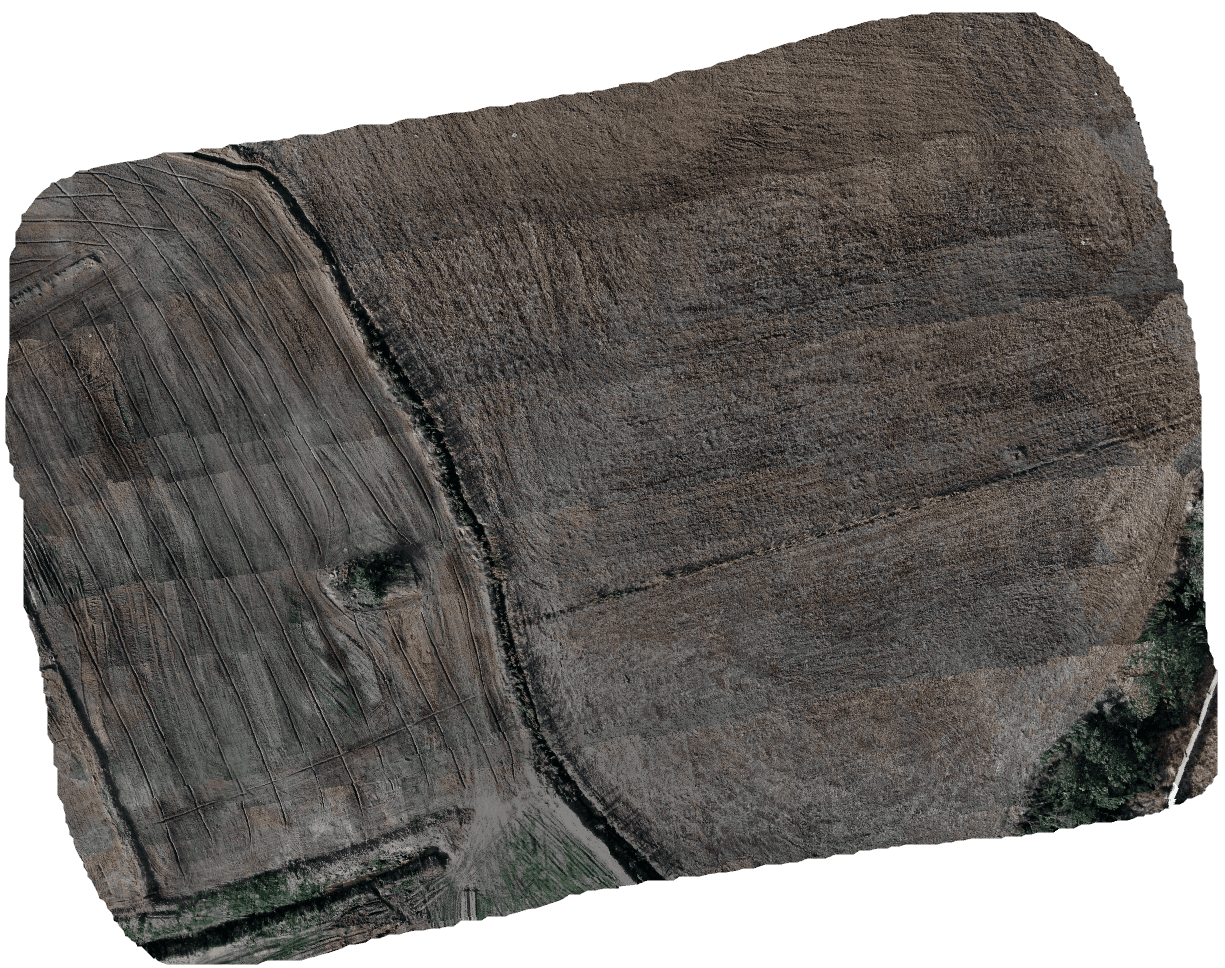}
        \caption{Our method}
        \label{figure6f}
    \end{subfigure}
    \caption{Orthoimages generated with different approaches}
    \label{figure6}
\end{figure*}

\subsection{Accuracy Evaluation}

In order to assess the accuracy of our reconstruction pipeline, we employed an unoptimized exhaustive matching method, which is known for its high computational cost but provides a reliable baseline for comparison. The resulting point clouds and trajectories were then aligned with the true GPS coordinates, which served as the ground truth for our evaluation. The accuracy of the trajectories generated by our system was quantitatively assessed using the evo tool~\cite{grupp2017evo}. As depicted in Fig.~\ref{figure4}, the trajectory obtained from our system exhibits a close alignment with the ground truth, thereby demonstrating the high accuracy of our reconstruction approach. This visual correspondence is further substantiated by the detailed absolute trajectory error (ATE) statistics presented in Table~\ref{table3}. These statistics provide a quantitative measure of the deviation between the estimated and true trajectories, confirming that our method achieves a high level of accuracy in reconstructing the scene.

\begin{table}[h]
\caption{ATE of translation and rotation}
\label{table3}
\begin{center}
\begin{tabular}{c|c c c|c c c}
\hline
\multirow{2}{*}{Scene}& \multicolumn{3}{c|}{Translation(m)}    & \multicolumn{3}{c}{Rotation(degree)}   \\ \cline{2-7}
                      & None.    & Opt.    & Opt.+Hom.  & None.    & Opt. & Opt.+Hom.    \\  \hline
Road                  & 0.95 & 0.81    & \cellcolor{red!15}0.68         & 1.03 & 1.00 & \cellcolor{red!15}0.84     \\ 
Hill                  & 2.61 & \cellcolor{red!15}1.37    & 1.41         & 1.68 & 1.56 & \cellcolor{red!15}1.40     \\  \hline
\end{tabular}
\end{center}
\end{table}

\subsection{System Integrity and Orthoimage Generation}

\begin{figure}[htb]
    \centering
    \begin{subfigure}[h]{0.47\columnwidth}
        \centering
        \includegraphics[width=\linewidth]{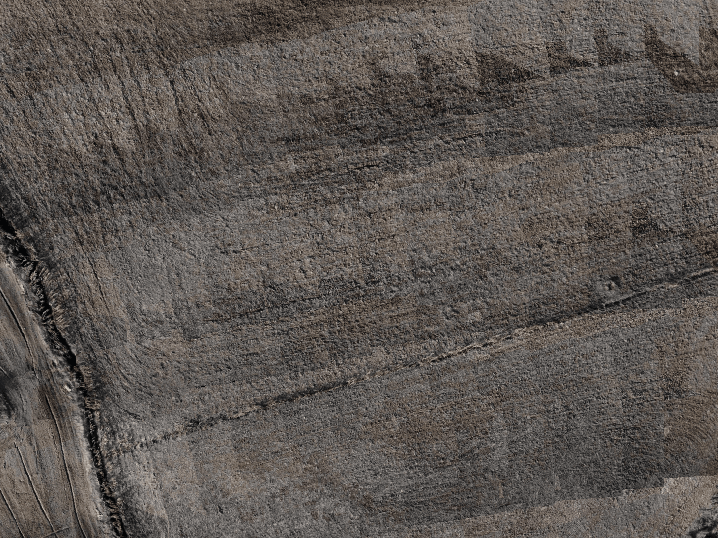}
        \caption{Without preprocessing}
        \label{figure7a}
    \end{subfigure}
    \begin{subfigure}[h]{0.47\columnwidth}
        \centering
        \includegraphics[width=\linewidth]{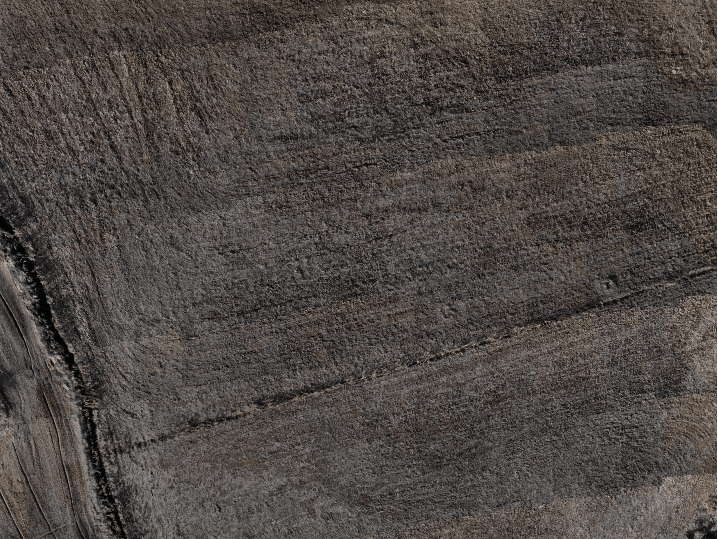}
        \caption{With preprocessing}
        \label{figure7b}
    \end{subfigure}
    \caption{Visual effects comparison of preprocessing}
    \label{figure7}
\end{figure}

Simplified terrain map is firstly generated with 4D millimeter-wave radar point clouds, as illustrated in Fig.~\ref{figure5}. The grid size is set to $1.0$, and the number of nearest neighbors used is $5$ during interpolation. Fig.~\ref{figure6} displays the orthoimages generated based on the terrain map using the proposed method, alongside the results obtained by processing our datasets with WebODM~\cite{OpenDroneMap} and Photoscan~\cite{verhoeven2011taking}. Table~\ref{table4} presents the computational time required for the entire orthophoto generation process using different methods. Our orthoimage generation employs an optimized BF matching algorithm, which is executed on the CPU for the most computationally intensive feature matching tasks. In contrast, WebODM and Photoscan leverage GPU acceleration and utilize prior GPS information to enhance their processing efficiency. Despite these advantages, our method achieves a 10× speedup compared to WebODM and a 3× speedup compared to Photoscan. Although our method exhibits minor deficiencies in visual appearance, it achieves comparable geospatial alignment accuracy to these commercial solutions. Additionally, Photoscan fails to generate the orthoimage in the upper-right corner for the hill scene, whereas our method maintains robustness in this scenario. The pre-processing steps, including illumination normalization and gaussian blur, have a significant impact on the visual quality of the generated images. Fig.~\ref{figure7} shows a portion of the hill scene to illustrate these effects. Orthoimages generated from preprocessed frames effectively eliminate seams caused by camera exposure. 

\section{CONCLUSIONS}

This paper introduces a novel rapid orthoimage generation approach utilizing multiple sensors. Compared to conventional methods that rely solely on cameras, the proposed approach achieves faster processing speeds and exhibits greater robustness in complex terrains, thereby meeting the demands of most practical applications. The optimized feature matching technique is versatile and well-integrated into our multi-sensor UAV system. This system has already been successfully deployed in agricultural applications, where it has demonstrated significant capability for enhancing precision monitoring. Although our system is capable of generating orthoimages well-aligned with geographic coordinates, further enhancement of the visual quality of the resultant images remains a potential area for future work. Moving forward, we will be committed to exploring solutions that further reduce the number of required feature points, aiming to accurately correct the camera pose with a small number of precisely selected feature points.

\begin{table}[h]
\caption{Comparison of computational efficiency. The unit of time is second.}
\label{table4}
\begin{center}
\begin{tabular}{c|c c c}
\hline
Scene      &WebODM~\cite{OpenDroneMap}    & Photoscan~\cite{verhoeven2011taking}  &Ours  \\ \hline
Road       & 490.533                      & 142.996                               &  \cellcolor{red!15}35.361     \\ 
Hill       & 927.676                      & 300.150                               &  \cellcolor{red!15}116.973     \\  \hline
\end{tabular}
\end{center}
\end{table}









\bibliographystyle{IEEEtran}
\bibliography{IEEEfull,mybibfile}

\begin{thebibliography}{10}
\providecommand{\url}[1]{#1}
\csname url@rmstyle\endcsname
\providecommand{\newblock}{\relax}
\providecommand{\bibinfo}[2]{#2}
\providecommand\BIBentrySTDinterwordspacing{\spaceskip=0pt\relax}
\providecommand\BIBentryALTinterwordstretchfactor{4}
\providecommand\BIBentryALTinterwordspacing{\spaceskip=\fontdimen2\font plus
\BIBentryALTinterwordstretchfactor\fontdimen3\font minus \fontdimen4\font\relax}
\providecommand\BIBforeignlanguage[2]{{%
\expandafter\ifx\csname l@#1\endcsname\relax
\typeout{** WARNING: IEEEtran.bst: No hyphenation pattern has been}%
\typeout{** loaded for the language `#1'. Using the pattern for}%
\typeout{** the default language instead.}%
\else
\language=\csname l@#1\endcsname
\fi
#2}}

\bibitem{mollick2023geospatial}
T.~Mollick, M.~G. Azam, and S.~Karim, ``Geospatial-based machine learning techniques for land use and land cover mapping using a high-resolution unmanned aerial vehicle image,'' \emph{Remote Sensing Applications: Society and Environment}, vol.~29, p. 100859, 2023.

\bibitem{baltsavias1996digital}
E.~P. Baltsavias, ``Digital ortho-images—a powerful tool for the extraction of spatial-and geo-information,'' \emph{ISPRS Journal of Photogrammetry and Remote sensing}, vol.~51, no.~2, pp. 63--77, 1996.

\bibitem{sheng2003true}
Y.~Sheng, P.~Gong, and G.~S. Biging, ``True orthoimage production for forested areas from large-scale aerial photographs,'' \emph{Photogrammetric Engineering \& Remote Sensing}, vol.~69, no.~3, pp. 259--266, 2003.

\bibitem{wu2020analysis}
K.-S. Wu, Y.-r. He, Q.-j. Chen, and Y.-m. Zheng, ``Analysis on the damage and recovery of typhoon disaster based on uav orthograph,'' \emph{Microelectronics Reliability}, vol. 107, p. 113337, 2020.

\bibitem{tiwari2020developing}
A.~Tiwari, M.~Silver, and A.~Karnieli, ``Developing object-based image procedures for classifying and characterising different protected agriculture structures using lidar and orthophoto,'' \emph{biosystems engineering}, vol. 198, pp. 91--104, 2020.

\bibitem{ullman1979interpretation}
S.~Ullman, ``The interpretation of structure from motion,'' \emph{Proceedings of the Royal Society of London. Series B. Biological Sciences}, vol. 203, no. 1153, pp. 405--426, 1979.

\bibitem{vallet2011photogrammetric}
J.~Vallet, F.~Panissod, C.~Strecha, and M.~Tracol, ``Photogrammetric performance of an ultra light weight swinglet uav,'' in \emph{UAV-g}, 2011.

\bibitem{verhoeven2011taking}
G.~Verhoeven, ``Taking computer vision aloft--archaeological three-dimensional reconstructions from aerial photographs with photoscan,'' \emph{Archaeological prospection}, vol.~18, no.~1, pp. 67--73, 2011.

\bibitem{moulon2013global}
P.~Moulon, P.~Monasse, and R.~Marlet, ``Global fusion of relative motions for robust, accurate and scalable structure from motion,'' in \emph{Proceedings of the IEEE international conference on computer vision}, 2013, pp. 3248--3255.

\bibitem{OpenDroneMap}
\BIBentryALTinterwordspacing
{Authors}, ``{ODM - A command line toolkit to generate maps, point clouds, 3D models and DEMs from drone, balloon or kite images},'' GitHub, 2020. [Online]. Available: \url{https://github.com/OpenDroneMap/ODM}
\BIBentrySTDinterwordspacing

\bibitem{bu2016map2dfusion}
S.~Bu, Y.~Zhao, G.~Wan, and Z.~Liu, ``Map2dfusion: Real-time incremental uav image mosaicing based on monocular slam,'' in \emph{2016 IEEE/RSJ International Conference on Intelligent Robots and Systems (IROS)}.\hskip 1em plus 0.5em minus 0.4em\relax IEEE, 2016, pp. 4564--4571.

\bibitem{zhang2005automatic}
L.~Zhang, \emph{Automatic digital surface model (DSM) generation from linear array images}.\hskip 1em plus 0.5em minus 0.4em\relax ETH Zurich, 2005.

\bibitem{pan2010region}
X.~Pan and S.~Lyu, ``Region duplication detection using image feature matching,'' \emph{IEEE Transactions on Information Forensics and Security}, vol.~5, no.~4, pp. 857--867, 2010.

\bibitem{hassaballah2016image}
M.~Hassaballah, A.~A. Abdelmgeid, and H.~A. Alshazly, ``Image features detection, description and matching,'' \emph{Image Feature Detectors and Descriptors: Foundations and Applications}, pp. 11--45, 2016.

\bibitem{garcia2010k}
V.~Garcia, E.~Debreuve, F.~Nielsen, and M.~Barlaud, ``K-nearest neighbor search: Fast gpu-based implementations and application to high-dimensional feature matching,'' in \emph{2010 IEEE International Conference on Image Processing}.\hskip 1em plus 0.5em minus 0.4em\relax IEEE, 2010, pp. 3757--3760.

\bibitem{sharma2016high}
K.~Sharma, ``High performance gpu based optimized feature matching for computer vision applications,'' \emph{Optik}, vol. 127, no.~3, pp. 1153--1159, 2016.

\bibitem{cornelis2008fast}
N.~Cornelis and L.~Van~Gool, ``Fast scale invariant feature detection and matching on programmable graphics hardware,'' in \emph{2008 IEEE Computer Society Conference on Computer Vision and Pattern Recognition Workshops}.\hskip 1em plus 0.5em minus 0.4em\relax IEEE, 2008, pp. 1--8.

\bibitem{vijayan2019flann}
V.~Vijayan and P.~Kp, ``Flann based matching with sift descriptors for drowsy features extraction,'' in \emph{2019 Fifth International Conference on Image Information Processing (ICIIP)}.\hskip 1em plus 0.5em minus 0.4em\relax IEEE, 2019, pp. 600--605.

\bibitem{muja2012fast}
M.~Muja and D.~G. Lowe, ``Fast matching of binary features,'' in \emph{2012 Ninth conference on computer and robot vision}.\hskip 1em plus 0.5em minus 0.4em\relax IEEE, 2012, pp. 404--410.

\bibitem{yang2022fast}
H.~Yang, Y.~Fu, D.~Chen, and Y.~Peng, ``A fast and effective panorama stitching algorithm on uav aerial images,'' in \emph{2022 14th International Conference on Computer Research and Development (ICCRD)}.\hskip 1em plus 0.5em minus 0.4em\relax IEEE, 2022, pp. 266--275.

\bibitem{szeliski2007image}
R.~Szeliski \emph{et~al.}, ``Image alignment and stitching: A tutorial,'' \emph{Foundations and Trends{\textregistered} in Computer Graphics and Vision}, vol.~2, no.~1, pp. 1--104, 2007.

\bibitem{adel2014image}
E.~Adel, M.~Elmogy, and H.~Elbakry, ``Image stitching based on feature extraction techniques: a survey,'' \emph{International Journal of Computer Applications}, vol.~99, no.~6, pp. 1--8, 2014.

\bibitem{wang2020review}
Z.~Wang and Z.~Yang, ``Review on image-stitching techniques,'' \emph{Multimedia Systems}, vol.~26, no.~4, pp. 413--430, 2020.

\bibitem{schonberger2016structure}
J.~L. Schonberger and J.-M. Frahm, ``Structure-from-motion revisited,'' in \emph{Proceedings of the IEEE conference on computer vision and pattern recognition}, 2016, pp. 4104--4113.

\bibitem{pan2024global}
L.~Pan, D.~Bar{\'a}th, M.~Pollefeys, and J.~L. Sch{\"o}nberger, ``Global structure-from-motion revisited,'' in \emph{European Conference on Computer Vision}.\hskip 1em plus 0.5em minus 0.4em\relax Springer, 2024, pp. 58--77.

\bibitem{davison2007monoslam}
A.~J. Davison, I.~D. Reid, N.~D. Molton, and O.~Stasse, ``Monoslam: Real-time single camera slam,'' \emph{IEEE transactions on pattern analysis and machine intelligence}, vol.~29, no.~6, pp. 1052--1067, 2007.

\bibitem{wang2019terrainfusion}
W.~Wang, Y.~Zhao, P.~Han, P.~Zhao, and S.~Bu, ``Terrainfusion: Real-time digital surface model reconstruction based on monocular slam,'' in \emph{2019 IEEE/RSJ International Conference on Intelligent Robots and Systems (IROS)}.\hskip 1em plus 0.5em minus 0.4em\relax IEEE, 2019, pp. 7895--7902.

\bibitem{snavely2006photo}
N.~Snavely, S.~M. Seitz, and R.~Szeliski, ``Photo tourism: exploring photo collections in 3d,'' in \emph{ACM siggraph 2006 papers}, 2006, pp. 835--846.

\bibitem{kerbl20233d}
B.~Kerbl, G.~Kopanas, T.~Leimk{\"u}hler, and G.~Drettakis, ``3d gaussian splatting for real-time radiance field rendering.'' \emph{ACM Trans. Graph.}, vol.~42, no.~4, pp. 139--1, 2023.

\bibitem{mildenhall2021nerf}
B.~Mildenhall, P.~P. Srinivasan, M.~Tancik, J.~T. Barron, R.~Ramamoorthi, and R.~Ng, ``Nerf: Representing scenes as neural radiance fields for view synthesis,'' \emph{Communications of the ACM}, vol.~65, no.~1, pp. 99--106, 2021.

\bibitem{wang2014automated}
H.~Wang, J.~Li, L.~Wang, H.~Guan, and Z.~Geng, ``Automated mosaicking of uav images based on sfm method,'' in \emph{2014 IEEE Geoscience and Remote Sensing Symposium}.\hskip 1em plus 0.5em minus 0.4em\relax IEEE, 2014, pp. 2633--2636.

\bibitem{lv2024fast}
J.~Lv, G.~Jiang, W.~Ding, and Z.~Zhao, ``Fast digital orthophoto generation: A comparative study of explicit and implicit methods,'' \emph{Remote Sensing}, vol.~16, no.~5, p. 786, 2024.

\bibitem{wu20153d}
Z.~Wu, S.~Song, A.~Khosla, F.~Yu, L.~Zhang, X.~Tang, and J.~Xiao, ``3d shapenets: A deep representation for volumetric shapes,'' in \emph{Proceedings of the IEEE conference on computer vision and pattern recognition}, 2015, pp. 1912--1920.

\bibitem{lao2018rolling}
Y.~Lao, O.~Ait-Aider, and A.~Bartoli, ``Rolling shutter pose and ego-motion estimation using shape-from-template,'' in \emph{Proceedings of the European Conference on Computer Vision (ECCV)}, 2018, pp. 466--482.

\bibitem{cui2017hsfm}
H.~Cui, X.~Gao, S.~Shen, and Z.~Hu, ``Hsfm: Hybrid structure-from-motion,'' in \emph{Proceedings of the IEEE conference on computer vision and pattern recognition}, 2017, pp. 1212--1221.

\bibitem{lynen2013robust}
S.~Lynen, M.~W. Achtelik, S.~Weiss, M.~Chli, and R.~Siegwart, ``A robust and modular multi-sensor fusion approach applied to mav navigation,'' in \emph{2013 IEEE/RSJ international conference on intelligent robots and systems}.\hskip 1em plus 0.5em minus 0.4em\relax IEEE, 2013, pp. 3923--3929.

\bibitem{falquez2016inertial}
J.~M. Falquez, M.~Kasper, and G.~Sibley, ``Inertial aided dense \& semi-dense methods for robust direct visual odometry,'' in \emph{2016 IEEE/RSJ International Conference on Intelligent Robots and Systems (IROS)}.\hskip 1em plus 0.5em minus 0.4em\relax IEEE, 2016, pp. 3601--3607.

\bibitem{hinzmann2018mapping}
T.~Hinzmann, J.~L. Sch{\"o}nberger, M.~Pollefeys, and R.~Siegwart, ``Mapping on the fly: Real-time 3d dense reconstruction, digital surface map and incremental orthomosaic generation for unmanned aerial vehicles,'' in \emph{Field and Service Robotics: Results of the 11th International Conference}.\hskip 1em plus 0.5em minus 0.4em\relax Springer, 2018, pp. 383--396.

\bibitem{he2025ligo}
D.~He, H.~Li, and J.~Yin, ``Ligo: A tightly coupled lidar-inertial-gnss odometry based on a hierarchy fusion framework for global localization with real-time mapping,'' \emph{IEEE Transactions on Robotics}, 2025.

\bibitem{han2019automated}
Y.~Han, J.~Choi, J.~Jung, A.~Chang, S.~Oh, and J.~Yeom, ``Automated coregistration of multisensor orthophotos generated from unmanned aerial vehicle platforms,'' \emph{Journal of Sensors}, vol. 2019, no.~1, p. 2962734, 2019.

\bibitem{chen2020densefusion}
L.~Chen, Y.~Zhao, S.~Xu, S.~Bu, P.~Han, and G.~Wan, ``Densefusion: Large-scale online dense pointcloud and dsm mapping for uavs,'' in \emph{2020 IEEE/RSJ International Conference on Intelligent Robots and Systems (IROS)}.\hskip 1em plus 0.5em minus 0.4em\relax IEEE, 2020, pp. 4766--4773.

\bibitem{hong2021radar}
Z.~Hong, Y.~Petillot, A.~Wallace, and S.~Wang, ``Radar slam: A robust slam system for all weather conditions,'' \emph{arXiv preprint arXiv:2104.05347}, 2021.

\bibitem{holder2019real}
M.~Holder, S.~Hellwig, and H.~Winner, ``Real-time pose graph slam based on radar,'' in \emph{2019 IEEE Intelligent Vehicles Symposium (IV)}.\hskip 1em plus 0.5em minus 0.4em\relax IEEE, 2019, pp. 1145--1151.

\bibitem{han20234d}
Z.~Han, J.~Wang, Z.~Xu, S.~Yang, L.~He, S.~Xu, J.~Wang, and K.~Li, ``4d millimeter-wave radar in autonomous driving: A survey,'' \emph{arXiv preprint arXiv:2306.04242}, 2023.

\bibitem{balamuruganky_ekf_2021}
\BIBentryALTinterwordspacing
Balamuruganky, ``Extended kalman filter (gps, velocity, and imu fusion),'' 2021, accessed: 2025-02-13. [Online]. Available: \url{https://github.com/balamuruganky/EKF_IMU_GPS}
\BIBentrySTDinterwordspacing

\bibitem{foley1987interpolation}
T.~A. Foley, ``Interpolation and approximation of 3-d and 4-d scattered data,'' \emph{Computers \& Mathematics with Applications}, vol.~13, no.~8, pp. 711--740, 1987.

\bibitem{grupp2017evo}
M.~Grupp, ``evo: Python package for the evaluation of odometry and slam.'' \url{https://github.com/MichaelGrupp/evo}, 2017.

\end{thebibliography}

\end{document}